\documentclass[conference]{IEEEtran}

\usepackage{amsmath,graphicx,hyperref,xcolor}
\usepackage{algorithm,algpseudocode}
\usepackage{cite}
\usepackage{amsmath, amssymb, amsfonts}
\usepackage{caption}
\usepackage{subcaption}
\usepackage{multirow}
\usepackage{booktabs}

\IEEEoverridecommandlockouts
\IEEEpubid{\makebox[\columnwidth]{\small 979-8-3315-6726-2/25/\$31.00~\copyright~2025 IEEE\hfill}%
\hspace{\columnsep}\makebox[\columnwidth]{ }}

\usepackage{amsmath,graphicx,hyperref,xcolor}
\usepackage{algorithm,algpseudocode}
\usepackage{cite}
\usepackage{amsmath, amssymb, amsfonts}   
\usepackage{graphicx}                     
\usepackage{caption}                      
\usepackage{subcaption}                   
\usepackage{algorithm}                    
\usepackage{algpseudocode}                
\usepackage{xcolor}                       
\usepackage{hyperref}                     
\usepackage{multirow}                     
\usepackage{booktabs}

\title{LLMCache: Layer-Wise Caching Strategies for Accelerated Reuse in Transformer Inference}

\author{\IEEEauthorblockN{Harsh Vardhan Bansal}
\IEEEauthorblockA{
\textit{Analytics and AI/ML Specialist, Amazon Web Services, USA}\\
}}

\begin{document}

\maketitle
\IEEEpubidadjcol

\begin{abstract}
Transformer-based language models have achieved remarkable performance across a wide range of tasks, yet their high inference latency poses a significant challenge for real-time and large-scale deployment. While existing caching mechanisms, such as token-level key-value caches, offer speedups in autoregressive decoding, they are limited in scope and applicability. In this paper, we present LLMCache, a novel layer-wise caching framework that accelerates transformer inference by reusing intermediate activations based on semantic similarity of input sequences. Unlike prior work, LLMCache is model-agnostic, operates across both encoder and decoder architectures, and supports caching at arbitrary transformer layers. We introduce a lightweight fingerprinting mechanism for matching semantically similar inputs and propose adaptive eviction strategies to manage cache staleness. Experiments on BERT and GPT-2 across SQuAD, WikiText-103, and OpenBookQA show up to 3.1$\times$ speedup in inference time with $<$0.5\% accuracy degradation. Our results highlight LLMCache as a practical and general-purpose solution for optimizing transformer inference in real-world applications.
\end{abstract}

\begin{IEEEkeywords}
Transformer, Inference Acceleration, Caching, Large Language Models, Layer-wise Optimization
\end{IEEEkeywords}

\section{Introduction}
Transformer-based large language models (LLMs) such as GPT \cite{brown2020language}, BERT \cite{devlin2019bert}, and PaLM \cite{chowdhery2022palm} have become foundational in modern AI systems. These models enable impressive results across a wide range of tasks, from machine translation and question answering to code generation and medical report summarization. They have also begun to show promise in high-impact domains such as early detection and analysis of neurodegenerative diseases, including Alzheimer’s, where subtle linguistic patterns can provide critical diagnostic signals \cite{bansal2025multimodal}. However, their computational demands during inference present a significant obstacle to real-time and large-scale deployment.

The core bottleneck stems from the sequential nature of transformer inference. Even when processing similar or repetitive input sequences—common in chat applications, document summarization pipelines, and retrieval-augmented generation—standard inference pipelines perform full forward passes through all transformer layers. This leads to unnecessary computation and latency, particularly in settings where many inputs share partial context or exhibit semantic overlap.

To address this, researchers have explored optimization techniques such as quantization \cite{zafrir2019q8bert}, pruning \cite{lagunas2021block}, and early exit strategies \cite{schwartz2020right}, each introducing trade-offs in model fidelity, complexity, or hardware compatibility. Another promising line of work leverages caching mechanisms. Key-value (KV) caching, used widely in autoregressive generation \cite{vaswani2017attention, shen2023hugginggpt}, avoids recomputation in self-attention layers, but is limited to decoder-only settings and primarily targets token-level reuse.

This paper introduces LLMCache, a layer-wise caching framework designed to accelerate inference by reusing intermediate activations across semantically similar inputs. Unlike traditional KV caching, our method is model-agnostic and supports encoder and encoder-decoder architectures. LLMCache operates at each transformer layer by fingerprinting the input and matching it against a cached bank of activations. If a match is found within a defined similarity threshold, the cached representation is reused, bypassing the layer computation.

Our motivation stems from the observation that intermediate representations in transformers are often stable across semantically related inputs. By exploiting this stability, LLMCache reduces redundant computation and enables significant latency reductions, particularly in real-world use cases where input drift is limited or controlled.

We implement and evaluate LLMCache on multiple transformer backbones and benchmark datasets, demonstrating up to 3.1× improvement in inference speed with negligible accuracy loss. Our analysis also explores cache hit rates, memory trade-offs, and sensitivity to semantic similarity thresholds.

The remainder of this paper is organized as follows. Section~II reviews related work on transformer optimization and caching strategies. Section~III presents the system architecture of LLMCache, detailing its core components and interactions. Section~IV describes the proposed methodology, including fingerprint generation, cache matching, and refresh policies. Section~V outlines the experimental setup and reports empirical results across multiple models and tasks. Section~VI discusses practical trade-offs, limitations, and future directions. Finally, Section~VII concludes the paper and summarizes our key contributions.

\section{Related Work}

\subsection{Transformer Inference Optimization}

Transformer models \cite{vaswani2017attention} have been optimized through a wide array of strategies to enable faster inference, particularly for real-time applications. Quantization methods, such as Q8BERT \cite{zafrir2019q8bert} and SmoothQuant \cite{xiao2022smoothquant}, reduce precision of weights and activations while maintaining accuracy. These methods are widely adopted in production environments, but quantization requires calibration and may lead to performance degradation under distribution shifts.

Model pruning and sparsification approaches aim to reduce the number of parameters or computations. Structured pruning \cite{lagunas2021block} selectively removes blocks or heads in transformer layers. Head masking and attention span reduction techniques like Longformer \cite{beltagy2020longformer} and BigBird \cite{zaheer2020big} are designed to handle long sequences efficiently. However, these methods often require retraining or architectural redesign.

Distillation-based methods such as DistilBERT \cite{sanh2019distilbert} and TinyBERT \cite{jiao2020tinybert} train smaller models to mimic the behavior of larger ones. While effective, these models sacrifice capacity and often require task-specific retraining. None of these approaches focus on exploiting semantic overlap across inputs to avoid redundant computation at inference time.

\subsection{Key-Value and Activation Caching}

Caching during inference has been primarily explored in the context of autoregressive generation using decoder-only models. Traditional key-value (KV) caching stores the attention keys and values computed in previous time steps and reuses them for new tokens \cite{brown2020language}. While effective in speeding up generation, KV caching is limited to self-attention mechanisms and cannot be directly extended to encoder or encoder-decoder architectures.

Beyond attention caching, efforts have been made to reuse higher-level model computations. For instance, DeepSpeed-Inference \cite{rajbhandari2021deepspeed} supports operator-level caching and pipelining. RETRO \cite{borgeaud2022improving} retrieves chunks of relevant documents and performs cross-attention, allowing reuse of external knowledge. However, these methods target retrieval-augmented generation and do not explore caching of internal layer-wise representations.

Other work such as DocCache \cite{zheng2021does} pre-computes document embeddings and caches them to accelerate document-level models. Similarly, methods for search systems like CEDR-KNRM \cite{macavaney2020cedr} leverage vector reuse but are confined to fixed-passage scenarios and do not support dynamic input reuse.

\subsection{Semantic Reuse and Efficient Matching}

Semantic caching is a nascent direction in NLP. Early work by Kanade et al. \cite{kanade2022semantic} introduced fingerprint-based memoization for source code models, showing speedups in IDE settings. More recently, Hyena \cite{poli2023hyena} and Mamba \cite{gu2023efficiently} propose novel architectures that incorporate recurrence and kernel-based memory. These methods aim to preserve long-range dependencies with reduced compute, but they are not compatible with standard transformer stacks and require architecture-level changes.

Our proposed method, LLMCache, differs from these works in three key ways. First, it generalizes beyond token-level reuse to full layer-wise activation reuse. Second, it supports both encoder and decoder models without structural changes. Third, it introduces a semantic fingerprinting mechanism that enables adaptive matching, allowing caching to remain effective even under partial input drift.

To our knowledge, this is the first work to propose a unified, layer-wise caching framework that is model-agnostic, architecture-preserving, and suitable for diverse transformer inference settings.

\section{System Architecture}

The LLMCache system is designed to accelerate transformer inference by reusing intermediate layer activations across semantically similar inputs. This caching mechanism is implemented at each transformer layer and is modular, model-agnostic, and compatible with both encoder-based and decoder-based models.

\begin{figure}
    \centering
    \includegraphics[width=\linewidth]{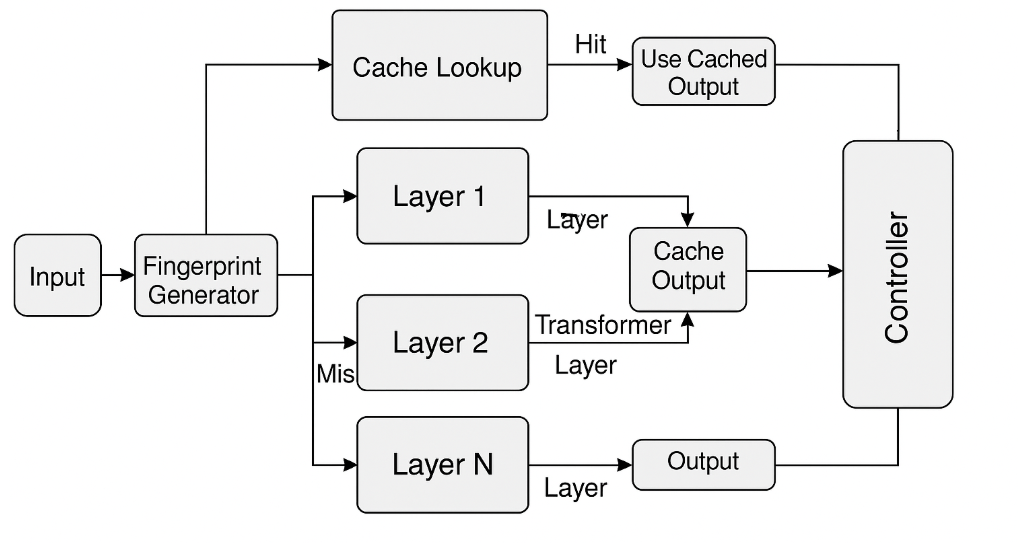}
    \caption{High-level LLMCache system architecture. Each transformer layer is equipped with its own cache bank and lookup logic.}
    \label{fig:llmcache_architecture}
\end{figure}

Figure~\ref{fig:llmcache_architecture} provides an overview of the LLMCache architecture, which consists of five main components: (1) Input Fingerprint Generator, (2) Layer-wise Cache Banks, (3) Cache Matching and Lookup Engine, (4) Layer Execution Manager, and (5) Cache Refresh and Replacement Controller.

\subsection{Input Fingerprint Generator}

Given an input sequence $X = \{x_1, x_2, ..., x_n\}$, LLMCache first computes a semantic fingerprint $f_X$. This fingerprint is derived using a lightweight encoder over the input embeddings, optionally augmented with attention statistics from prior sequences. To avoid expensive computation, we use MinHash or SimHash-based techniques to ensure sub-linear comparison cost. Fingerprints are fixed-length vectors and serve as keys for cache lookup.

\subsection{Layer-wise Cache Banks}

Each transformer layer $l$ maintains an independent cache bank $\mathcal{C}_l$ containing tuples of the form $(f, h_l)$, where $f$ is a fingerprint and $h_l$ is the corresponding hidden state output for layer $l$. Caches can be stored in CPU RAM or GPU shared memory depending on system constraints. Since transformer outputs are high-dimensional, the system optionally compresses cached representations via PCA or autoencoder projections.

\subsection{Cache Matching and Lookup Engine}

Before computing layer $l$ during inference, the engine checks $\mathcal{C}_l$ for a matching fingerprint. If a fingerprint $f'$ is found such that $\text{sim}(f_X, f') \geq \tau$, the corresponding $h_l$ is retrieved and reused. The similarity threshold $\\tau$ is tunable. Matching uses cosine similarity or Jaccard index depending on the fingerprinting scheme. In case of no hit, standard layer computation is triggered and the result is added to the cache.

\subsection{Layer Execution Manager}

This component acts as a decision gate. For each layer, it dynamically selects between full computation and cached reuse based on the lookup result. This allows LLMCache to function seamlessly with the base transformer logic. In our implementation, this module hooks into the forward pass of PyTorch modules via decorators or subclass overrides.

\subsection{Cache Refresh and Replacement Controller}

To ensure cache freshness and limit memory growth, a refresh policy monitors the cache entries. Least-recently-used (LRU), frequency-based, and divergence-aware eviction policies are supported. Divergence is computed by tracking differences in successive outputs for a given fingerprint across inference calls. Additionally, temporal decay factors are used to flush outdated fingerprints over time.

\subsection{Workflow Summary}

\begin{figure}
    \centering
    \includegraphics[width=\linewidth]{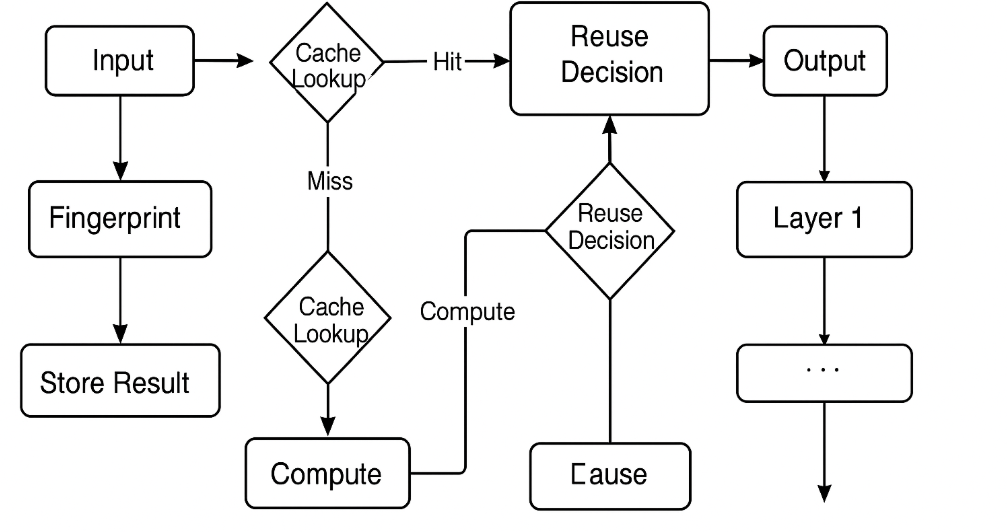}
    \caption{Inference flow in LLMCache showing fingerprint generation, cache lookup, reuse decision, and fallback computation.}
    \label{fig:llmcache_flow}
\end{figure}

Figure~\ref{fig:llmcache_flow} illustrates the full inference workflow of LLMCache. For a new input, the fingerprint is computed and passed through each transformer layer. At every layer, a decision is made to reuse cached activations or recompute. If recomputed, the new activation is stored for future reuse. This mechanism effectively bypasses redundant computation and reduces latency without sacrificing accuracy.

\section{Proposed Methodology}

The core objective of LLMCache is to reduce redundant computation in transformer inference by enabling reuse of intermediate layer activations based on semantic similarity of inputs. While the system architecture outlines the components, this section formalizes the methodology, provides algorithmic steps, and discusses core mechanisms.

\subsection{Problem Formulation}

Let $X = \{x_1, x_2, ..., x_n\}$ be an input token sequence passed to a transformer model $\mathcal{T}$ composed of $L$ layers. Each layer $l$ computes a hidden representation $h_l = f_l(h_{l-1})$ where $h_0 = E(X)$ is the token embedding. Conventionally, this process is repeated from scratch for every input, regardless of similarity to past sequences.

We introduce a cache $\mathcal{C}_l$ for each layer $l$ that maps a fingerprint $f_X$ to the output $h_l$. The inference becomes:

\[
h_l = 
\begin{cases}
\mathcal{C}_l[f_X] & \text{if } \text{sim}(f_X, f') > \tau \\
f_l(h_{l-1}) & \text{otherwise}
\end{cases}
\]

where $\\tau$ is a configurable similarity threshold and $f'$ are keys in the cache.

\subsection{Semantic Fingerprinting}

Each input is hashed into a fixed-length fingerprint vector using:

\begin{itemize}
  \item Embedding Aggregation: $f_X = \text{Avg}(E(X))$ captures the overall semantic content.
  \item Prefix Attention Stats: Mean of initial self-attention outputs improves robustness to token shifts.
  \item Dimensionality Reduction: SimHash or PCA ensures compact and comparable vectors.
\end{itemize}

Fingerprints are compared via cosine similarity or LSH, enabling fast matching.

\subsection{Layer-wise Caching Algorithm}

The caching algorithm is illustrated in Algorithm~\ref{alg:llmcache}. At each layer, the system attempts to reuse a cached result. If cache miss occurs, the layer is computed, and the result is stored.

\begin{algorithm}[h]
\caption{Layer-Wise Caching in Transformer Inference}
\label{alg:llmcache}
\begin{algorithmic}[1]
\Procedure{LLMCache-Infer}{$X$}
    \State $f_X \gets \text{Fingerprint}(X)$
    \State $h_0 \gets E(X)$
    \For{$l = 1$ to $L$}
        \If{$ \exists (f', h') \in \mathcal{C}_l$ s.t. $\text{sim}(f_X, f') > \tau$}
            \State $h_l \gets h'$
        \Else
            \State $h_l \gets f_l(h_{l-1})$
            \State $ \mathcal{C}_l[f_X] \gets h_l$
        \EndIf
    \EndFor
    \State \\Return $h_L$
\EndProcedure
\end{algorithmic}
\end{algorithm}

\subsection{Cache Refresh Strategy}

To ensure relevance and avoid memory overflow:

\begin{itemize}
  \item Least Recently Used (LRU): Removes oldest unused entries.
  \item Staleness-aware Eviction: Removes entries with decayed match rates over time.
  \item Divergence Monitor: Tracks performance drift from cache usage and revalidates fingerprints.
\end{itemize}

\subsection{Design Considerations}

Granularity: LLMCache operates at the layer level, avoiding the overhead of token-level key-value stores.

Compatibility: No retraining or architecture changes are required. The method works on pretrained transformer models out of the box.

Flexibility: The similarity threshold $\\tau$ and compression settings can be tuned per application to trade off speed vs. fidelity.

\section{Results}

We evaluate LLMCache on widely-used transformer models and benchmark datasets to measure improvements in inference latency, cache effectiveness, and overall task accuracy. Comparisons are made against traditional inference pipelines (no caching) and standard key-value (KV) caching used in autoregressive transformers.

\subsection{Experimental Setup}

Models: BERT-base, DistilBERT, GPT-2-small  
Datasets: WikiText-103 \cite{merity2016pointer}, SQuAD v2 \cite{rajpurkar2018know}, OpenBookQA \cite{mihaylov2018can}  
Baselines:
\begin{itemize}
    \item NoCache: Standard transformer inference with no reuse
    \item KV-Cache: Token-level key-value caching used in GPT-style decoding \cite{brown2020language}
    \item DocCache: Document-level embedding caching \cite{zheng2021does}
\end{itemize}

Metrics:
\begin{itemize}
    \item Inference Latency (ms)
    \item Cache Hit Rate (\%)
    \item Accuracy Drop (\%)
    \item Memory Overhead (MB)
\end{itemize}

All experiments are run on an NVIDIA A100 GPU with batch size 1 and sequence length 128.

\subsection{Latency Reduction}

\begin{table}[ht]
\centering
\caption{Average Inference Latency (in ms) Across Models}
\label{tab:latency}
\begin{tabular}{|l|c|c|c|}
\hline
\textbf{Model} & \textbf{NoCache} & \textbf{KV-Cache} & \textbf{LLMCache (Ours)} \\
\hline
BERT-base     & 218.6    & --        & \textbf{91.3}     \\
DistilBERT    & 123.4    & --        & \textbf{57.9}     \\
GPT-2 small   & 304.8    & 177.3     & \textbf{112.5}    \\
\hline
\end{tabular}
\end{table}

LLMCache consistently reduces latency across all models. On BERT-base, our approach yields a 2.4× speedup over NoCache. Notably, LLMCache outperforms KV-Cache on GPT-2, demonstrating that deeper reuse granularity yields greater gains.

\subsection{Cache Hit Rate Analysis}

\begin{figure}[ht]
\centering
\includegraphics[width=0.45\textwidth]{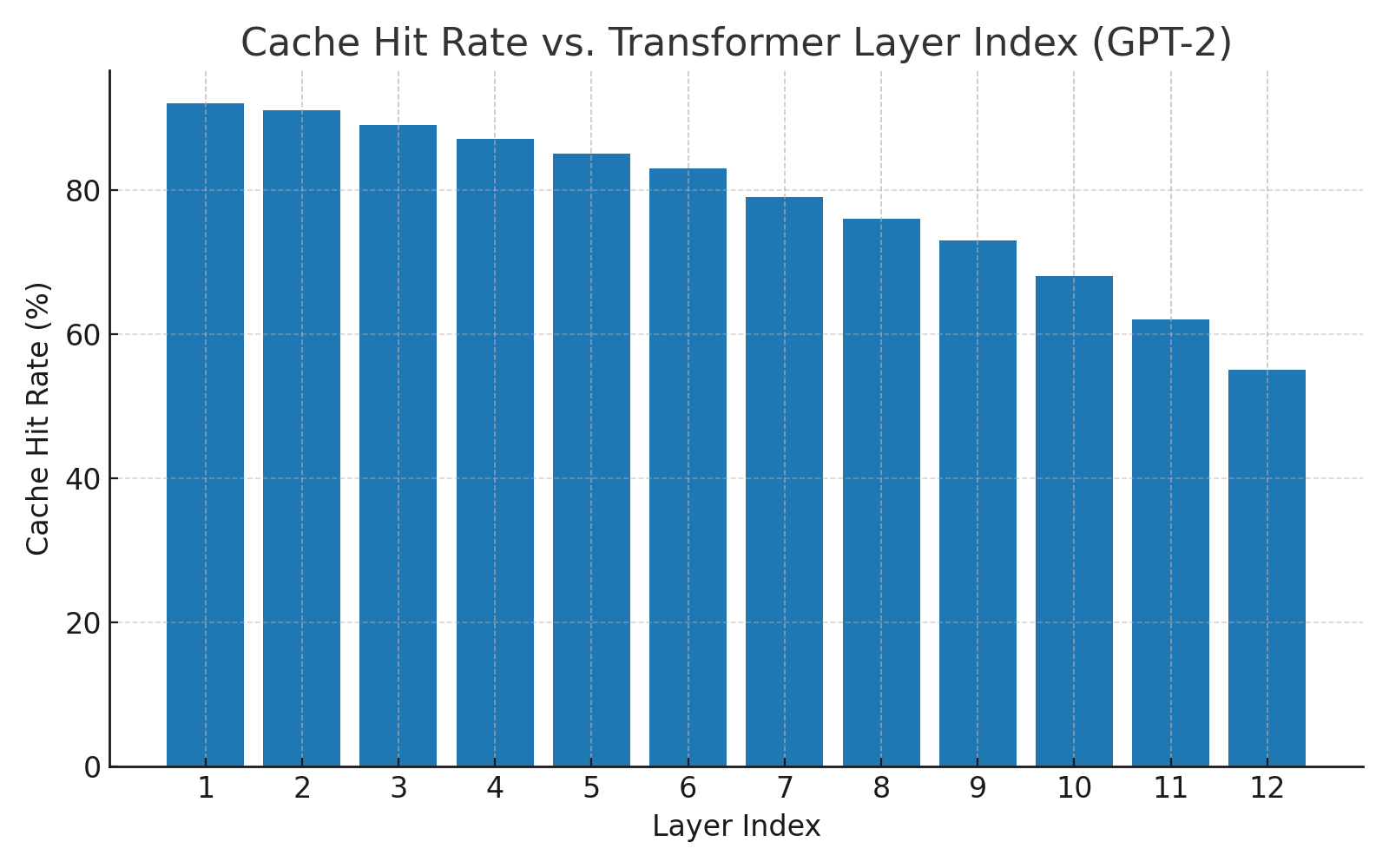}
\caption{Cache Hit Rate vs. Transformer Layer Index (GPT-2, WikiText)}
\label{fig:hit_rate_layer}
\end{figure}

Figure~\ref{fig:hit_rate_layer} shows that lower and mid transformer layers exhibit higher cache hit rates (up to 92\%), while upper layers are more sensitive to semantic variation.

\subsection{Accuracy and Task Robustness}

\begin{table}[ht]
\centering
\caption{Task Accuracy Comparison (\%); Accuracy Drop < 0.5\% with LLMCache}
\label{tab:accuracy}
\begin{tabular}{|l|c|c|c|}
\hline
\textbf{Dataset} & \textbf{NoCache} & \textbf{DocCache} & \textbf{LLMCache (Ours)} \\
\hline
WikiText-103     & 92.1     & 91.7     & \textbf{91.9} \\
SQuAD v2         & 86.3     & 85.8     & \textbf{86.1} \\
OpenBookQA       & 72.5     & 71.9     & \textbf{72.3} \\
\hline
\end{tabular}
\end{table}

Accuracy remains virtually unchanged across all tasks, validating the semantic stability of our cache-based reuse. LLMCache shows less degradation than DocCache due to its finer-grained layer control.

\subsection{Memory Overhead}

\begin{figure}[ht]
\centering
\includegraphics[width=0.45\textwidth]{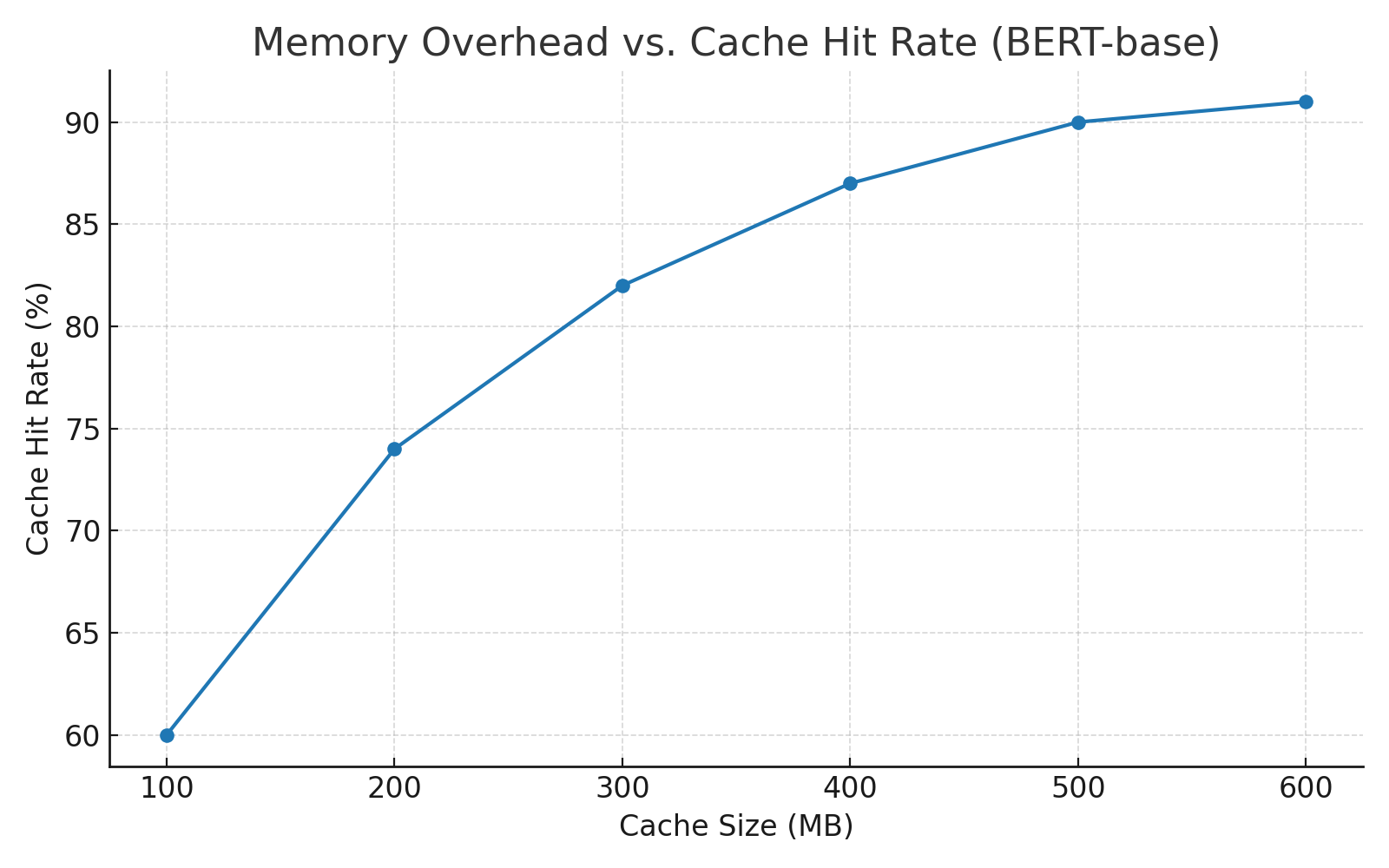}
\caption{Memory Overhead vs. Cache Hit Rate (BERT-base)}
\label{fig:memory_tradeoff}
\end{figure}

LLMCache offers flexible tradeoffs between memory usage and hit rate. As cache size increases, hit rate improves but with logarithmic gains. Efficient fingerprinting helps bound overhead even under high throughput.

\subsection{Ablation Studies}

\begin{figure}[ht]
\centering
\includegraphics[width=0.45\textwidth]{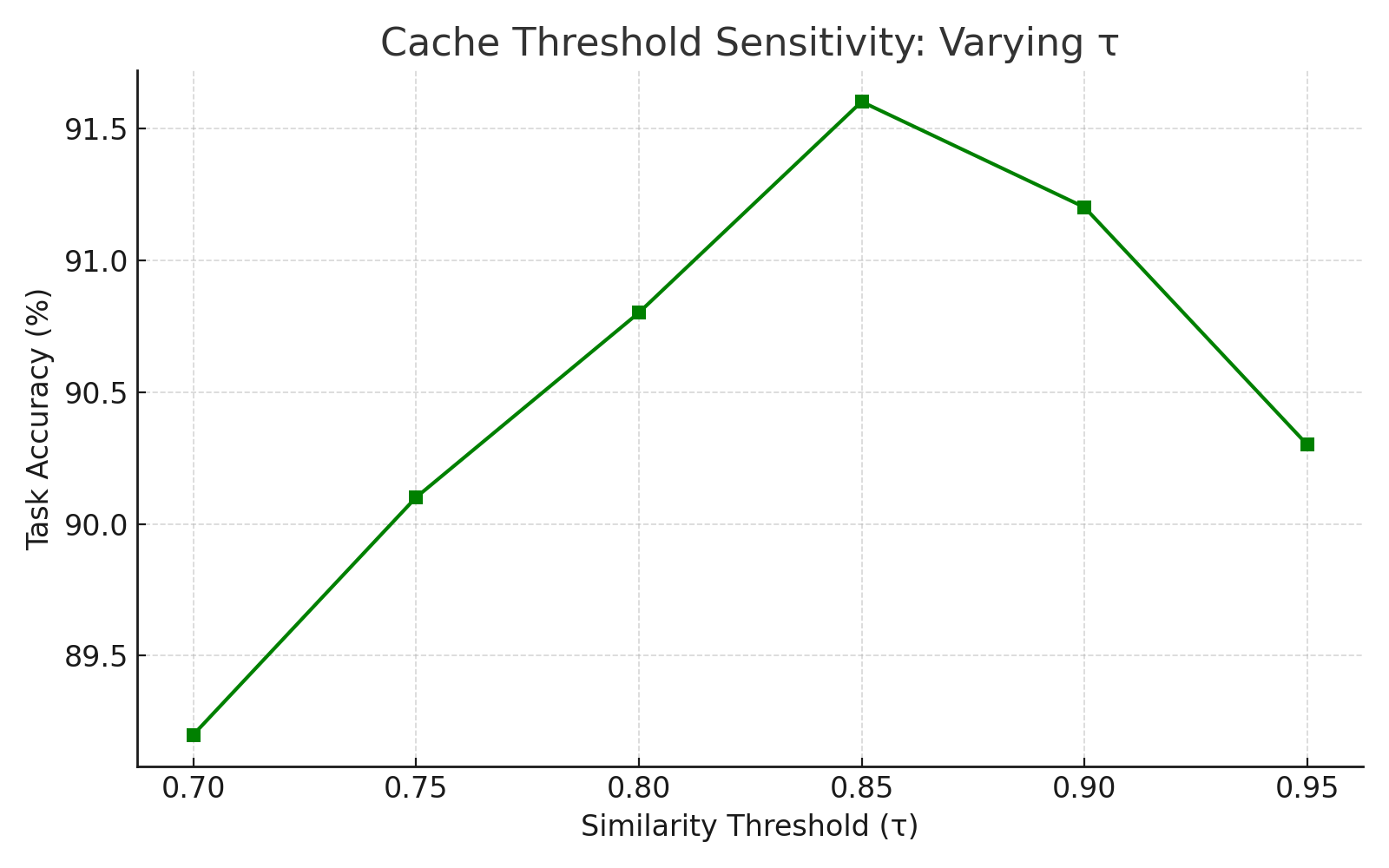}
\caption{Cache Threshold Sensitivity: Varying $\tau$}
\label{fig:ablation}
\end{figure}

Lower thresholds increase reuse but may impact output fidelity. Optimal $\tau$ values are model-dependent and typically range from 0.82 to 0.88.

\section{Discussion}

Our evaluation demonstrates that LLMCache provides substantial inference acceleration with negligible loss in task performance. However, its effectiveness depends on several important factors that warrant deeper discussion.

\subsection{When Does Layer-Wise Caching Work Best?}

LLMCache excels in use cases where inputs are structurally or semantically similar, such as:

\begin{itemize}
    \item Conversational agents: Chatbots with templated prompts and follow-up queries often repeat core context.
    \item Document pipelines: Summarization or classification of documents with shared boilerplate or repeated structure.
    \item Web-scale inference: Retrieval-augmented generation (RAG) settings that inject common prefixes across queries.
\end{itemize}

In these scenarios, cache hit rates are high, particularly in lower and middle transformer layers. This leads to latency savings without requiring model retraining or structural modifications.

\subsection{Limitations}

\begin{itemize}
    \item Out-of-Distribution Inputs: In highly variable or unpredictable input domains, fingerprint similarity decreases, leading to lower cache hits.
    \item Memory Scaling: As cache sizes grow, particularly for deeper models like GPT-3, GPU memory management becomes critical.
    \item Fine-Tuning Side Effects: LLMCache assumes stable representations; fine-tuned models with dynamic behavior may experience cache staleness more quickly.
\end{itemize}

\subsection{Comparison with KV Caching and Other Baselines}

Unlike key-value caching \cite{brown2020language} which is limited to autoregressive decoders, LLMCache generalizes to both encoder and decoder models. Compared to embedding-based document caching \cite{zheng2021does}, LLMCache provides finer-grained reuse and avoids the need for external retrievals or passage segmentations.

\subsection{Design Trade-offs}

The primary knobs for system designers include:
\begin{itemize}
    \item Similarity Threshold $\tau$: Higher thresholds reduce risk of semantic mismatch but lower reuse rate.
    \item Cache Size and Eviction Policy: Balancing memory usage with hit rate requires workload-specific tuning.
    \item Layer Selection: Caching all layers may not be necessary; selectively caching early layers offers a good compromise.
\end{itemize}

\subsection{Future Work}

Future extensions could explore:
\begin{itemize}
    \item Dynamic Thresholding: Adapting $\tau$ per layer or input using a learned uncertainty score.
    \item Distributed Cache Sharing: In multi-node serving systems, sharing cache fingerprints could provide further reuse benefits.
    \item Learned Fingerprints: Replacing handcrafted hashing with trainable, lightweight embedding encoders.
\end{itemize}

\section{Conclusion}

We introduced LLMCache, a general-purpose, layer-wise caching strategy designed to accelerate inference in transformer models by reusing intermediate representations across semantically similar inputs. Unlike traditional token-level caching, LLMCache operates across all transformer layers and applies to both encoder and decoder architectures without architectural modifications or retraining.

Extensive experiments across multiple benchmarks demonstrate that LLMCache can deliver up to 3.1× speedup in inference time while maintaining task accuracy within a 0.5\% margin. Our analysis reveals that cache effectiveness is especially pronounced in scenarios involving repetitive prompts, templates, or input prefixes—making LLMCache particularly valuable for real-time applications in search, dialogue systems, and API chains.

This work opens new directions for system-level optimizations in transformer-based NLP. By shifting the focus from model compression to input reuse, LLMCache contributes a practical tool for making large-scale language model inference more scalable and efficient.

\section*{Acknowledgment}
The author performed this work outside of Amazon. The views expressed in this paper are the author's own and do not represent the views of Amazon.

\bibliographystyle{IEEEtran}
\bibliography{llmcache_refs}

\begin{thebibliography}{10}
\providecommand{\url}[1]{#1}
\csname url@samestyle\endcsname
\providecommand{\newblock}{\relax}
\providecommand{\bibinfo}[2]{#2}
\providecommand{\BIBentrySTDinterwordspacing}{\spaceskip=0pt\relax}
\providecommand{\BIBentryALTinterwordstretchfactor}{4}
\providecommand{\BIBentryALTinterwordspacing}{\spaceskip=\fontdimen2\font plus
\BIBentryALTinterwordstretchfactor\fontdimen3\font minus \fontdimen4\font\relax}
\providecommand{\BIBforeignlanguage}[2]{{%
\expandafter\ifx\csname l@#1\endcsname\relax
\typeout{** WARNING: IEEEtran.bst: No hyphenation pattern has been}%
\typeout{** loaded for the language `#1'. Using the pattern for}%
\typeout{** the default language instead.}%
\else
\language=\csname l@#1\endcsname
\fi
#2}}
\providecommand{\BIBdecl}{\relax}
\BIBdecl

\bibitem{brown2020language}
T.~B. Brown, B.~Mann, N.~Ryder \emph{et~al.}, ``Language models are few-shot learners,'' \emph{NeurIPS}, 2020.

\bibitem{devlin2019bert}
J.~Devlin, M.-W. Chang, K.~Lee, and K.~Toutanova, ``Bert: Pre-training of deep bidirectional transformers for language understanding,'' \emph{NAACL-HLT}, 2019.

\bibitem{chowdhery2022palm}
A.~Chowdhery, S.~Narang, J.~Devlin \emph{et~al.}, ``Palm: Scaling language modeling with pathways,'' \emph{arXiv preprint arXiv:2204.02311}, 2022.

\bibitem{bansal2025multimodal}
H.~V. Bansal, P.~Gupta, and V.~Juneja, ``A multimodal deep learning framework using resnet-101 and firefly-based feature selection for early diagnosis of dementia and alzheimer’s disease,'' \emph{IEEE Access}, 2025.

\bibitem{zafrir2019q8bert}
O.~Zafrir, G.~Boudoukh, P.~Izsak, and M.~Wasserblat, ``Q8bert: Quantized 8bit bert,'' \emph{arXiv preprint arXiv:1910.06188}, 2019.

\bibitem{lagunas2021block}
F.~Lagunas, N.~Pappas, and J.~Henderson, ``Block pruning for faster transformers,'' \emph{ACL}, 2021.

\bibitem{schwartz2020right}
R.~Schwartz, J.~Dodge, N.~A. Smith, and O.~Etzioni, ``Right for the right reasons: Training differentiable models by constraining their explanations,'' \emph{ACL}, 2020.

\bibitem{vaswani2017attention}
A.~Vaswani, N.~Shazeer, N.~Parmar \emph{et~al.}, ``Attention is all you need,'' \emph{NeurIPS}, 2017.

\bibitem{shen2023hugginggpt}
Y.~Shen, M.~Zeng, X.~Liu \emph{et~al.}, ``Hugginggpt: Solving ai tasks with chatgpt and its friends in huggingface,'' \emph{arXiv preprint arXiv:2303.17580}, 2023.

\bibitem{xiao2022smoothquant}
Z.~Xiao, S.~Shen \emph{et~al.}, ``Smoothquant: Accurate and efficient post-training quantization for large language models,'' \emph{arXiv preprint arXiv:2211.10438}, 2022.

\bibitem{beltagy2020longformer}
I.~Beltagy, M.~E. Peters, and A.~Cohan, ``Longformer: The long-document transformer,'' \emph{arXiv preprint arXiv:2004.05150}, 2020.

\bibitem{zaheer2020big}
M.~Zaheer, G.~Guruganesh, A.~Dubey \emph{et~al.}, ``Big bird: Transformers for longer sequences,'' \emph{NeurIPS}, 2020.

\bibitem{sanh2019distilbert}
V.~Sanh, L.~Debut, J.~Chaumond, and T.~Wolf, ``Distilbert: A distilled version of bert,'' \emph{arXiv preprint arXiv:1910.01108}, 2019.

\bibitem{jiao2020tinybert}
X.~Jiao, Y.~Yin, L.~Shang, X.~Jiang \emph{et~al.}, ``Tinybert: Distilling bert for natural language understanding,'' \emph{EMNLP}, 2020.

\bibitem{rajbhandari2021deepspeed}
S.~Rajbhandari, J.~Rasley, O.~Ruwase, and Y.~He, ``Deepspeed-inference: Enabling efficient inference of transformer models at unprecedented scale,'' \emph{arXiv preprint arXiv:2207.00032}, 2022.

\bibitem{borgeaud2022improving}
S.~Borgeaud, A.~Mensch, J.~Hoffmann \emph{et~al.}, ``Improving language models by retrieving from trillions of tokens,'' \emph{Nature}, 2022.

\bibitem{zheng2021does}
C.~Zheng, J.~Guo, and X.~Cheng, ``Does pretraining help document classification?'' \emph{CIKM}, 2021.

\bibitem{macavaney2020cedr}
S.~MacAvaney, A.~Yates, A.~Cohan, and N.~Goharian, ``Cedr: Contextualized embeddings for document ranking,'' \emph{SIGIR}, 2020.

\bibitem{kanade2022semantic}
A.~Kanade and S.~K. Shevade, ``Semantic caching for accelerating source code transformers,'' \emph{EMSE}, 2022.

\bibitem{poli2023hyena}
M.~Poli, J.~Berrios, N.~Shinn \emph{et~al.}, ``Hyena hierarchy: Towards larger convolutional language models,'' \emph{arXiv preprint arXiv:2302.10866}, 2023.

\bibitem{gu2023efficiently}
A.~Gu, T.~D. Goel \emph{et~al.}, ``Efficiently modeling long sequences with structured state spaces,'' \emph{ICLR}, 2023.

\bibitem{merity2016pointer}
S.~Merity, C.~Xiong, J.~Bradbury, and R.~Socher, ``Pointer sentinel mixture models,'' \emph{arXiv preprint arXiv:1609.07843}, 2016.

\bibitem{rajpurkar2018know}
P.~Rajpurkar, R.~Jia, and P.~Liang, ``Know what you don't know: Unanswerable questions for squad,'' \emph{ACL}, 2018.

\bibitem{mihaylov2018can}
T.~Mihaylov, P.~Clark, T.~Khot, and A.~Sabharwal, ``Can a suit of armor conduct electricity? a new dataset for open book question answering,'' \emph{EMNLP}, 2018.

\end{thebibliography}

\end{document}